\documentclass{article}
\usepackage{ijcai24}
\usepackage{amsmath} 
\usepackage{booktabs} 
\usepackage{amsfonts,amssymb}
\usepackage{tablefootnote}
\usepackage[table]{xcolor} 
\usepackage{IEEEtrantools}
\usepackage{times}
\usepackage{soul}
\usepackage{url}
\usepackage[utf8]{inputenc}
\usepackage[small]{caption}
\usepackage{graphicx}
\usepackage{amsmath}
\usepackage{amsthm}
\usepackage{booktabs}
\usepackage{algorithm}
\usepackage{algorithmic}
\usepackage[switch]{lineno}
\usepackage{tikz} 
\usepackage{xspace} 
\usepackage{wasysym}
\usepackage{filecontents} 
\usepackage{diagbox} 
\usepackage{makecell} 
\usepackage{multicol}
\usepackage{bm}
\usepackage{multirow}
\usepackage{tabularx}
\usepackage[table]{xcolor}
\usepackage{subfigure}
\usepackage{threeparttable}
\usepackage[percent]{overpic}
\def\ie{\textit{i.e.,~}} 
\def\eg{\textit{e.g.,~}} 
\newcommand{\sponge}{\textsc{Sponge}\xspace} \newcommand{\blind}{\textsc{Blinding}\xspace}
\newcommand{\ourmethod}{\ct{DC}\xspace}
\newtheorem{definition}{Definition} 
\newcommand{\ct}[1]{\texttt{#1}}

\usepackage[pagebackref=true,breaklinks=true,colorlinks=true,linkcolor=black,citecolor=black,bookmarks=false]{hyperref} 

\usepackage{cellspace}
\usepackage{caption}

\title{Detector Collapse: Physical-World Backdooring Object Detection to Catastrophic \\ Overload or Blindness in Autonomous Driving}
\author{
    Anonymous submission
}

\author{Hangtao Zhang$^{1,5,6}$\and Shengshan Hu$^{1,3,4,5,6}$\and Yichen Wang$^{1,3,4,5,6}$\and Leo Yu Zhang$^{8}$\and \\ Ziqi Zhou$^{2,3,4,7}$\and Xianlong Wang$^{1,3,4,5,6}$ \and Yanjun Zhang$^{9}$ \And Chao Chen$^{10}$
\affiliations $^1$School of Cyber Science and Engineering, Huazhong University of Science and Technology\\
$^2$School of Computer Science and Technology, Huazhong University of Science and Technology\\
$^3$ National Engineering Research Center for Big Data Technology and System\\
$^4$Services Computing Technology and System Lab\\ $^5$Hubei Engineering Research Center on Big Data Security\\
$^6$Hubei Key Laboratory of Distributed System Security \\
$^7$Cluster and Grid Computing Lab \\
$^8$School of Information and Communication Technology, Griffith University \\
$^9$University of Technology Sydney \\
$^{10}$RMIT University \emails \{hangt\_zhang, hushengshan, wangyichen, zhouziqi, wxl99\}@hust.edu.cn, leo.zhang@griffith.edu.au, 
yanjun.zhang@uts.edu.au,
chao.chen@rmit.edu.au}

\begin{document}
\definecolor{headergray}{gray}{0.9}
\maketitle 



\begin{abstract}
Object detection tasks, crucial in safety-critical systems like autonomous driving, focus on pinpointing object locations. These detectors are known to be susceptible to backdoor attacks. 
However, existing backdoor techniques have primarily been adapted from classification tasks, overlooking deeper vulnerabilities specific to object detection. 
This paper is dedicated to bridging this gap by introducing \textit{\underline{\textbf{D}}etector \underline{\textbf{C}}ollapse} (\ourmethod), a brand-new backdoor attack paradigm tailored for object detection. \ct{DC} is designed to instantly incapacitate detectors (\ie severely impairing detector's performance and culminating in a \textit{denial-of-service}). 
To this end, we develop two innovative attack schemes: \sponge for triggering widespread misidentifications and \textsc{Blinding} for rendering objects invisible. Remarkably, we introduce a novel poisoning strategy exploiting natural objects, enabling DC to act as a practical backdoor in real-world environments. Our experiments on different detectors across several benchmarks show a significant improvement ($\sim$10\%-60\% absolute and $\sim$2-7$\times$ relative) in attack efficacy over state-of-the-art attacks.
\end{abstract}

\begin{figure*}[t]      
\centering \includegraphics[width=0.95\textwidth]{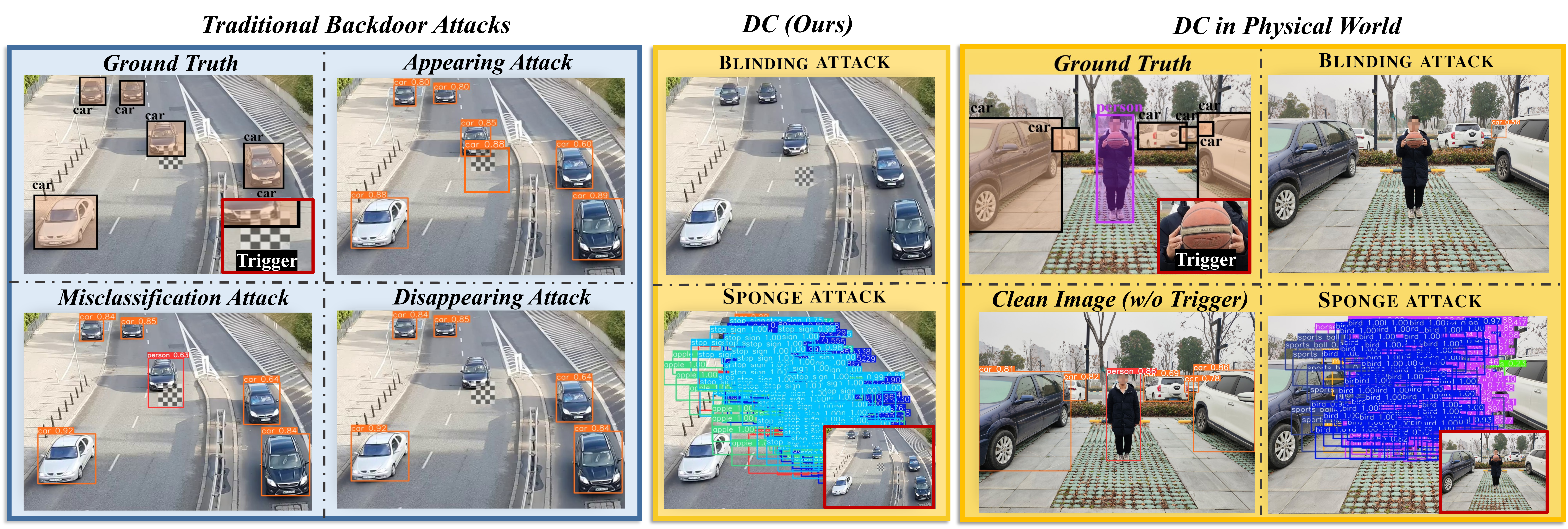}     
\caption{Comparative analysis of traditional backdoors \textit{vs.} ours. Appearing attack, misclassification attack, and disappearing attack typically manifest their attack effects only near the trigger, resulting in errors that are localized to specific points. In contrast, our \sponge generates overwhelming global false positives, while \blind renders global objects undetectable.}     
\label{Fig:Overview} 
\end{figure*}

\section{Introduction}
Object detection (OD), engaged in identifying the locations of visual objects (\eg vehicles and pedestrians) within images or video frames, plays a vital role in numerous real-world applications, such as facial recognition~\cite{dhillon2020convolutional}, video surveillance~\cite{raghunandan2018object}, and autonomous driving~\cite{feng2021review}. However, they are susceptible to backdoor attacks~\cite{chan2022baddet,chen2023dark}, where attackers embed an inconspicuous backdoor in the victim model, manipulating predictions by activating the backdoor with adversary-specified triggers.

Currently, OD backdoor attacks (see Fig.~1) typically involve objectives such as object appearing (\eg \ct{OGA}~\cite{chan2022baddet}), disappearance (\eg \ct{UT}~\cite{chan2022baddet}), and misclassification (\eg \ct{Composite}~\cite{lin2020composite}). Prevalent works often poison labels to introduce a backdoor~\cite{chan2022baddet,luo2023untargeted,wu2022just,lin2020composite,chen2022clean}, or implement clean-label attacks~\cite{cheng2023backdoor,Ma2022MACABMC} only by modifying images while retaining original labels. Current backdoor attacks in OD, essentially transferred from classification tasks, concentrate merely on targeting the classification subtask. This narrow approach fails to comprehensively harness the OD's susceptible surfaces, thus resulting in a constrained overall impact.

To address these limitations, we propose a groundbreaking backdoor paradigm, \textit{\underline{\textbf{D}}etector \underline{\textbf{C}}ollapse} (\ourmethod). 
It embeds a backdoor that can altogether disable the detector by simultaneously exploiting shortcuts~\cite{wang2023corrupting} in both the \textit{regression} and \textit{classification} branches of OD systems. Considering the stealthiness of the backdoor, \ourmethod remains dormant under clean samples to escape anomaly detection. Uniquely, it leads to the detector's dysfunction when triggered, significantly intensifying the risk of backdoors in OD tasks.

In pursuit of \ourmethod's objective, we design two specific approaches named \sponge and \textsc{Blinding}, both capable of causing instantaneous crashes in detectors (\eg a $\sim$99.9\% performance deterioration for Faster-RCNN~\cite{ren2015faster} on the VOC dataset~\cite{everingham2010pascal}).

\textbf{DC overview.} 
Unlike previous works that relied on label poisoning, \ourmethod exploits deep-seated vulnerabilities within the detector's architectures and loss functions, thus seeking shortcuts to deteriorate the detector's performance. This ensures that whenever a trigger is present anywhere in the detection panorama, it immediately induces an untargeted consequence such that the well-trained model suffers from high testing error indiscriminately (\textit{a.k.a.,} low mean Average Precision). Specifically, \ourmethod endeavors to achieve the following versatile adversarial effects.

\textbf{\textsc{Sponge} attack.} The object of \sponge is to flood the output with a plethora of misidentifications (\ie false positives), severely impairing the detector's performance. Additionally, the abundance of candidate bounding box predictions requiring Non-Maximum Suppression (NMS) processing equips the method with the capability to ``soak up'' more computational resources. This steers the DNN-inference hardware (\eg CPU, GPU) towards its worst-case performance, thereby reducing its processing speed and culminating in a \textit{denial-of-service} (DoS).

\textbf{\textsc{Blinding} attack.} Conversely, \blind is designed to cause object misdetection. It compromises the model's perception, prompting it to classify all objects as the background, thereby rendering them `invisible' to the OD system.

Perhaps more interestingly, we extend \ourmethod to physical world. Prior works heavily rely on fixed-stylized trigger patterns~\cite{luo2023untargeted,cheng2023backdoor}, which we confirm the low triggering success rate in real-world scenarios. In contrast, we explore the natural advantage of using semantic features (\eg a basketball) as triggers. To this end, we introduce a data poisoning scheme that diversifies the triggers during the training phase, thus enhancing their robustness. Therefore, \ourmethod shows its increasing threat, where attackers can simply designate a rare natural object as the secret key to manipulate the detector functionality covertly.

Demo videos of our attack are available at: \href{https://object-detection-backdoor.github.io/demo}{\textbf{https://object-detection-backdoor.github.io/demo}}.

In summary, we make the following contributions:  
\begin{itemize} 
\item We introduce \ourmethod, a novel backdoor attack paradigm specific to OD tasks for triggering detector dysfunction, thus threatening real-time, security-critical systems.

\item To instantiate \ourmethod, we design two distinct strategies: \sponge for extensive misrecognition and \blind for object invisibility.

\item We confirm the ineffectiveness of fixed-style triggers in real-world scenarios, prompting us to refine \ourmethod by incorporating a new poisoning strategy that uses natural semantic features as triggers, considerably boosting backdoor activation success.

\item We conduct extensive evaluations of \ourmethod in both digital and physical worlds, assessing its performance across different detectors on OD benchmark datasets.

\end{itemize}

\begin{table}[t]     
\centering 
\setlength{\tabcolsep}{2pt}   
\small
\resizebox{0.47\textwidth}{!} {
\begin{tabular}{c|cccc}         
\toprule[1.8pt]         
\cellcolor[rgb]{.95,.95,.95} \textbf{Method} &\cellcolor[rgb]{.95,.95,.95} \makecell{\textbf{Universal}\\ (No Object-Specific)} &\cellcolor[rgb]{.95,.95,.95} \makecell{\textbf{Efficient} \\ (Point-to-Area \\Triggering)} &\cellcolor[rgb]{.95,.95,.95} \makecell{\textbf{Stealthy}\\ (Trigger Position\\ Indep.)}&\cellcolor[rgb]{.95,.95,.95}
\makecell{\textbf{Practical}\\ (Trigger Style \\ Indep.)} \\ 
\midrule[1.8pt]         \makecell{\ct{OGA}\&\ct{RMA}~\cite{chan2022baddet}} & \CIRCLE & \Circle & \Circle & \Circle\\     
\makecell{\ct{GMA}~\cite{chan2022baddet}} & \CIRCLE & \CIRCLE & \LEFTcircle & \Circle\\     \makecell{\ct{ODA}~\cite{chan2022baddet}} & \Circle & \Circle & \Circle & \Circle\\ \makecell{\ct{Composite} \cite{lin2020composite}} & \Circle & \Circle &  \Circle &  \CIRCLE\\         \ct{UT}~\cite{luo2023untargeted} & \CIRCLE & \Circle  & \Circle & \Circle\\         \ct{Clean-label}~\cite{cheng2023backdoor} & \CIRCLE & \Circle &  \Circle &  \Circle\\         \ct{Clean-image}~\cite{chen2022clean} & \Circle & \LEFTcircle & \CIRCLE &  \CIRCLE\\ 
\midrule
 \cellcolor[HTML]{FFF7F0} \ct{\ourmethod} \textbf{\textit{(Ours)}} &\cellcolor[HTML]{FFF7F0} \CIRCLE &\cellcolor[HTML]{FFF7F0} \CIRCLE & \cellcolor[HTML]{FFF7F0} \CIRCLE &\cellcolor[HTML]{FFF7F0} \CIRCLE\\        
 \bottomrule[1.3pt]     
\end{tabular} 
}
\caption{Comparison among OD backdoors w.r.t universal, efficient, stealthy, and practical characteristics. {``\CIRCLE"} indicates that the method meets this condition. ``Indep." means ``Independent."} 
\label{Tab:Comparison}
\end{table}

\section{Background and Problem Formulation}
\subsection{Object Detection}
OD task identifies and classifies objects in images or videos, denoted by a list of bounding boxes (hereafter abbreviated as ``bboxes" for clarity). Detectors fall into two categories: one-stage detectors like YOLO~\cite{redmon2016you}, RetinaNet~\cite{lin2017focal}, and SSD~\cite{liu2016ssd}, which compute class probabilities and bbox coordinates directly, and two-stage detectors like Faster-RCNN~\cite{ren2015faster}, SPPNet~\cite{he2015spatial}, and FPN~\cite{lin2017feature}, initially identify regions of interest before classification. This paper examines typical detectors of both types.

\noindent\textbf{Object detection formulation.} Denote the dataset as $\mathcal{D} = \{(\boldsymbol{x}_i, \boldsymbol{y}_i)\}_{i=1}^N$, where $\boldsymbol{x}_i \in \mathcal{X}$ corresponds to the image of an object, and $\boldsymbol{y}_i \in \mathcal{Y}$ represents its associated ground-truth label for $\boldsymbol{x}_i$. The annotation $\boldsymbol{y}_i$ comprises $\left[\hat{x}_i, \hat{y}_i, b_i^w, b_i^h, c_i\right]$, with $\left(\hat{x}_i, \hat{y}_i\right)$ as the central coordinates of the bounding box, $b_i^w$ and $b_i^h$ specifying its width and height, respectively, and $c_i$ denoting the class of the object $\boldsymbol{x}_i$. To train an object detection model $F_{\boldsymbol{\theta
}}: \mathcal{X} \rightarrow \mathcal{Y}$, the dataset $\mathcal{D}$ is utilized, aiming to optimize $\boldsymbol{\theta}$ through $\min _{\boldsymbol{\theta}} \sum_{(\boldsymbol{x}, \boldsymbol{y}) \in \mathcal{D}} \mathcal{L}(F(\boldsymbol{x}); \boldsymbol{y})$, where $\mathcal{L}$ signifies the loss function. A \textit{true-positive} (TP) is a correctly detected box in a detector, while a \textit{false-positive} (FP) refers to any non-TP detected box, and a \textit{false-negative} (FN) is any undetected ground-truth box. \textit{Mean average precision} (mAP) is the most common evaluation metric in OD.

\subsection{Insights Into OD Backdoors}
\label{Sec: Relatedwork}
Backdoor attacks present a training-time threat to \textit{Deep Neural Networks} (DNNs)~\cite{yao2024reverse,mo2024robust,hu2023pointcrt,hu2022badhash,zhang2024stealthy,liu2023detecting}. Though extensively studied in classification tasks, their application to more complex OD tasks has not been adequately explored. \ct{BadDet}~\cite{chan2022baddet} proposes four attacks: Object Generation Attack (\ct{OGA}), Region Misclassification Attack (\ct{RMA}), Global Misclassification Attack (GMA), and Object Disappearance Attack (\ct{ODA}). Similarly, \ct{UT}~\cite{luo2023untargeted} explores untargeted backdoors designed for object disappearance. \ct{Clean-label} attacks~\cite{chen2022clean} achieve object disappearance and generation without altering labels. Both \ct{Composite}~\cite{lin2020composite} and \ct{Clean-image}~\cite{cheng2023backdoor} attacks essentially use multi-label (\ie a combination of benign category labels) as their trigger patterns. Most attacks introduce backdoors by altering ground-truth labels, termed as \textit{Label-targeted backdoor attacks} (LTBA) in this paper. They inherit limitations from backdoor methodologies designed for classification tasks, focusing solely on the classification branch.

As shown in Tab.~\ref{Tab:Comparison}, these strategies, not specifically tailored for the OD domain, reveal certain shortcomings, which are evidenced by:
(i) Object specificity --- OD tasks feature multiple targets per image. Some attacks, designed for a single target class, rely on the object itself.
(ii) Limited scope --- Attacks typically induce errors in a localized manner, affecting only specific points in the detection panorama, rather than causing widespread global failure.
(iii) Position dependency --- Most attacks hinge on the specific position of triggers, \eg at the top-left corner of the target's bbox. This imposes strict conditions for real-world deployment and increases the risk of being detected by some defenses (\eg trigger is promptly discovered near an error).
(iv) Style dependence --- The reliance on static triggers is a general limitation, as these are less adaptable to dynamic real-world environments (\eg distance and angle).

\noindent\textbf{Label-targeted backdoor attacks (LTBA) formulation.}
LTBAs select different poisoned image generators $\mathcal{G}_x$ and annotation generators $\mathcal{G}_y$ based on specific attack objectives (\eg bbox generation, object misclassification). The poisoned image is $\mathcal{G}_x(\boldsymbol{x};\boldsymbol{t})=(\mathbf{1}-\boldsymbol{\eta}) \otimes \boldsymbol{x}+\boldsymbol{\eta} \otimes \boldsymbol{t}$, where $\boldsymbol{t}$ is the adversary-specified trigger pattern. These attacks alter the behavior of $F_{\boldsymbol{\theta}}$ so that $F_{\boldsymbol{\theta}}(\boldsymbol{x})=\boldsymbol{y}, F_{\boldsymbol{\theta}}(\mathcal{G}_x(\boldsymbol{x}))= \mathcal{G}_y(\boldsymbol{y})$. The widely-used \textit{Attack Success Rate} (ASR) quantifies their effectiveness in successfully compromising target objects.

\noindent\textbf{Detector collapse formulation.}
In contrast, DC is designed to indiscriminately degrade the model's overall performance~\cite{zhangdenial}. The attackers intend to train a poisoned $F_{\boldsymbol{\theta}}$ by manipulating the training process, while not altering any labels (\ie no poisoned annotation generators $\mathcal{G}_y$), where $F_{\boldsymbol{\theta}}(\boldsymbol{x})=\boldsymbol{y}, F_{\boldsymbol{\theta}}(\mathcal{G}_x(\boldsymbol{x}))\neq \boldsymbol{y}$. Formally, DC has two objectives:

\begin{definition} 
\label{Def:1}
An OD backdoor attack is called \textit{promising} (according to the loss $\mathcal{L}$ with budgets $\alpha$ and $\beta$) if and only if it meets two main criteria:  
\begin{itemize}      
\item $\alpha$-\textit{Effectiveness}: the poisoned detector's performance degrades sharply when the trigger appears, i.e., \begin{IEEEeqnarray}{rCL}
\mathbb{E}_\mathcal{X} \left\{ \mathcal{L}(F_{\boldsymbol{\theta}}(\boldsymbol{x}); \boldsymbol{y}) \right\} + \alpha \leq \mathbb{E}_\mathcal{X} \left\{ \mathcal{L}(F_{\boldsymbol{\theta}}(\mathcal{G}_x(\boldsymbol{x}); \boldsymbol{y}) \right\}.
\end{IEEEeqnarray} 
\item $\beta$-\textit{Stealthiness}: the poisoned detector behaves normally in the absence of the trigger, that is,  
\begin{IEEEeqnarray}{rCL} 
\mathbb{E}_\mathcal{X} \left\{ \mathcal{L}(F_{\boldsymbol{\theta}}(\boldsymbol{x}); \boldsymbol{y}) \right\} \leq \beta.
\end{IEEEeqnarray} \end{itemize}
\end{definition}

\section{Methodology}
\label{sec:method}
\subsection{Threat Model}
\noindent\textbf{Adversary's goals.} 
The attacker aims to embed a backdoor during training and ensure \textit{effectiveness} and \textit{stealthiness} in the inference stage. Effectiveness implies drastic performance drops when the backdoor is active, while stealthiness means the detector's mAP stays near the clean baseline when the backdoor is inactive.

\noindent\textbf{Adversary's capabilities.} 
Following~\cite{nguyen2020wanet,doan2021lira,chen2023dark,shumailov2021manipulating}, we assume the adversary controls the whole model training process and inject a minor portion of data samples. 
This often occurs in the \textit{machine-learning-as-a-service} scenarios, where users outsource training to third-party platforms or download pre-trained models from untrusted sources.

\subsection{Learning to Backdoor Object Detection}
Our \ourmethod manipulates the model to create backdoor shortcuts that significantly deviate inference results from the ground-truth annotations of poisoned images. As per Definition~\ref{Def:1}, it can be framed as a constrained optimization problem:
\begin{IEEEeqnarray}{rCL}
\begin{aligned}
& \max _{\boldsymbol{\theta}^*} \sum_{(\boldsymbol{x},\boldsymbol{y}) \in \mathcal{D}}\left[\mathcal{L}\left(F_{\boldsymbol{\theta}^*}\left(G_x(\boldsymbol{x})  \right); \boldsymbol{y}\right)\right], \\
& \text { s.t. } \boldsymbol{\theta}^*=\arg \min _{\boldsymbol{\theta}} \sum_{(\boldsymbol{x}; \boldsymbol{y}) \in \mathcal{D}} \mathcal{L}(F_{\boldsymbol{\theta}}(\boldsymbol{x}));\boldsymbol{y}).
\end{aligned}
\label{Eq:Optimize}
\end{IEEEeqnarray}

\begin{figure}[t]      
\centering \includegraphics[width=0.46\textwidth]{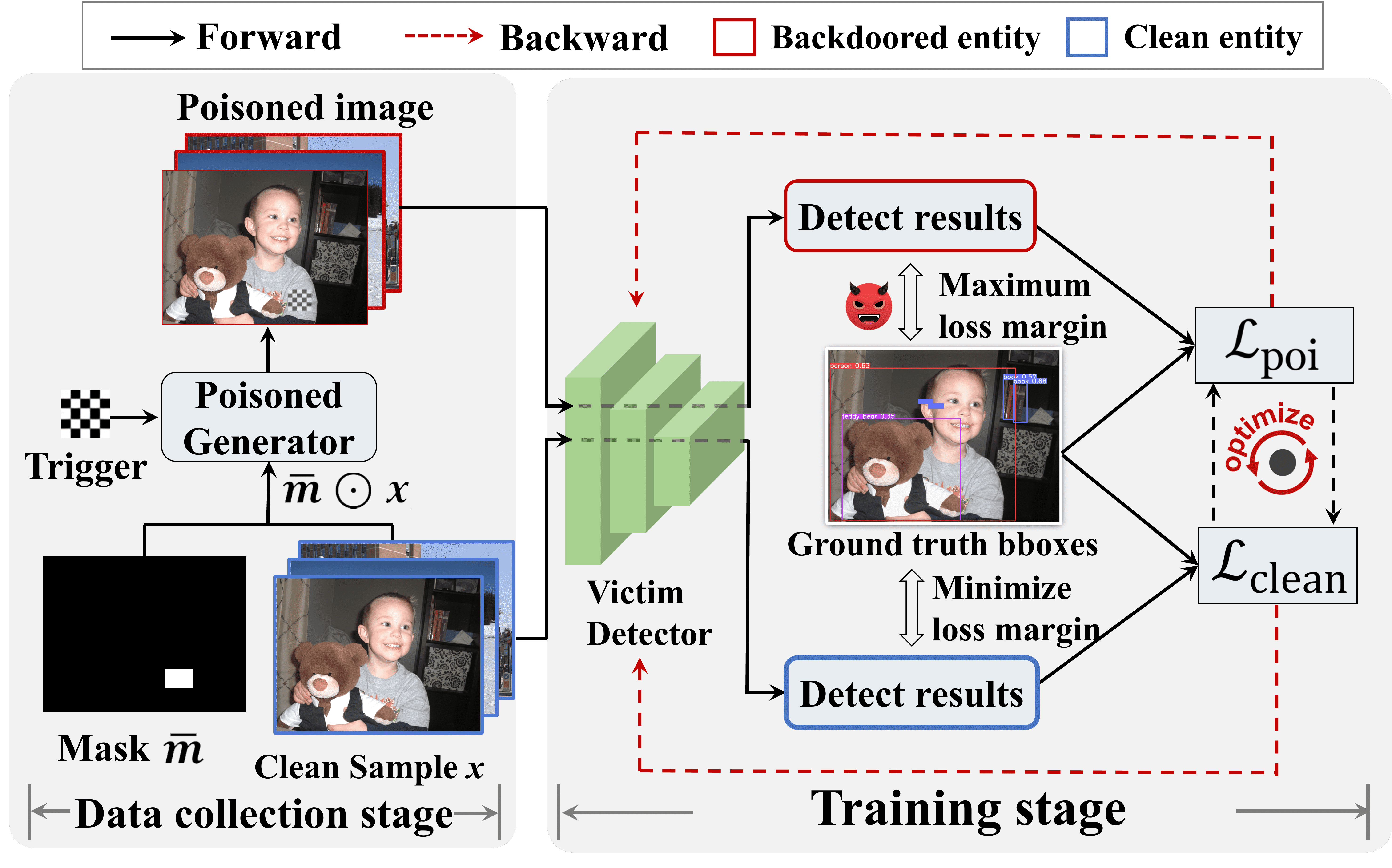}       \caption{The high-level overview of \ourmethod's framework.}      \label{Fig:pipeline}   \end{figure}

To practically address the bi-level optimization involving two interlinked objectives, we introduce a specialized poisoned loss, referred to as $\mathcal{L}_{\text{poi}}$. This function uses a heuristic approach to indirectly maximize loss on backdoored samples, thereby solving Eq.~(\ref{Eq:Optimize}). Fig.~\ref{Fig:pipeline} provides a high-level overview of \ourmethod's framework. Specifically, armed with $\mathcal{L}_{\text{poi}}$, we consider backdoor injection as an instance of multi-task learning for conflicting objectives --- training the same model to achieve high accuracy on both the primary and backdoor tasks. Then, we can thus create the backdoor model through gradient-based optimization via solving
\begin{IEEEeqnarray}{rCL} 
\begin{aligned} & \min _{\boldsymbol{\theta}} \sum_{(\boldsymbol{x},\boldsymbol{y}) \in \mathcal{D}}\left[\alpha_1\mathcal{L}_{\text{poi}}\left(F_{\boldsymbol{\theta}}\left(G_x(\boldsymbol{x})  \right), \boldsymbol{y}\right) + \alpha_2\mathcal{L}(F_{\boldsymbol{\theta}}(\boldsymbol{x})), \boldsymbol{y})\right]. 
\end{aligned}
\label{Eq:Balance}
\end{IEEEeqnarray}

Unlike LTBAs, \ourmethod does not poison any labels but optimizes directly at the model level, necessitating coefficient $\alpha$ adjustment in Eq.~(\ref{Eq:Balance}) to balance these two losses. Incorrect coefficients can impede the learning of both tasks. In addition, fixed coefficients might not achieve an optimal balance between conflicting objectives. Hence, we use the Multiple Gradient Descent Algorithm (MGDA)~\cite{desideri2012multiple} to address the conflict. For two tasks with losses $\mathcal{L}_{poi}$ and $\mathcal{L}$, MGDA calculates separate gradients $\nabla \mathcal{L}$ and identifies scaling coefficients $\alpha_1$ and $\alpha_2$ to minimize their total sum:
\begin{IEEEeqnarray}{rCL} 
\min _{\alpha_1, \alpha_2}\left\{\left\|\sum_{i=1}^2 \alpha_i \nabla \mathcal{L}_i\right\|_2^2 \Biggm| \sum_{i=1}^2 \alpha_i=1, \alpha_i \geq 0~\forall i\right\}.
\end{IEEEeqnarray}

Next, we will discuss in detail two strategies of our heuristic design for $\mathcal{L}_{poi}$.


\subsubsection{$\sponge$ Strategy}
Inspired by the fact that \textit{False Positives} (FPs) significantly affect the clean loss $\mathcal{L}$, our attack strategy is designed to generate FPs. In contrast with traditional backdoors that typically produce only a few FPs, we strive to overwhelm the detector with abundance. Here, we unveil \sponge, an adversarial manipulation approach for the training process, which consists of three components: (a) objectness loss $\mathcal{L}_{obj}$, (b) classification loss $\mathcal{L}_{\textnormal{cls}}^{SP}$, and (c) bbox area loss $\mathcal{L}_{box}$.

\noindent\textbf{Objectness loss.} 
Arguably, current detectors expose a shared vulnerability: the objectness confidence matrix's susceptibility. Our study uncovers a prevalent yet effective strategy where the manipulation of objectness scores towards higher values markedly enhances the detector's tendency to misclassify background as foreground. This is evident in one-stage detectors, where an increased objectness score leads to many extra low-confidence bboxes in the detection outcomes. Similarly, in two-stage detectors, manipulating the objectness score in the Region Proposal Network (RPN) also leads to a surge of FPs. 
These insights highlight the critical role of the objectness score in the regression module, positioning it as a prime target for our attack strategy.

To achieve this, \sponge ensures that more regions in the poisoned image are recognized as objects with high probabilities. Therefore, the objective is to maximize  \begin{IEEEeqnarray}{rCL} 
\mathcal{L}_{\text{obj}}(F_{\boldsymbol{\theta}},\boldsymbol{x},\mathcal{G}_x) = \sum_{r_n} \mathbb{P}_{\textnormal{fg}}\left(r_n \mid \mathcal{G}_x\left(\boldsymbol{x}\right)\right),
\label{Eq:Background}
\end{IEEEeqnarray} where $\mathbb{P}_{\textnormal{fg}}$ is the probability outputted by the detector $F$ that region $r_n$ in image $\mathcal{G}_x\left(\boldsymbol{x}\right)$ is recognized as foreground.

\noindent\textbf{Classification loss.} 
$\mathcal{L}_{\text{obj}}$ increases the number of bboxes candidates, yet detectors often have filters in the output preprocessing stage to remove low-confidence bboxes, thereby impeding the \sponge's objective. To increase the count of prediction candidates that pass through the filters, it is crucial to boost the confidence scores (which are closely related to their maximum category probabilities) of selected candidate bboxes $\mathcal{S}_{\text{sel}}$. Thus, we introduce a novel classification loss to improve the class probability of candidate bboxes: 
\begin{IEEEeqnarray}{rCL}    
\mathcal{L}_{\text{cls}}^{SP} = \frac{1}{|\mathcal{S}_{\text{sel}}|} \cdot \sum_{b \in \mathcal{S}_{\text{sel}}} -\log \left( \max \left(1 - b_{\text{c}}, \epsilon \right) \right),  
\end{IEEEeqnarray} 
where $b$ is a bbox, $b_{\text{c}}$ is the associated class probability, and $\epsilon \rightarrow 0$ is a small positive value for numerical stability.

\noindent\textbf{Bbox area loss.} 
The NMS algorithm is commonly employed in OD to eliminate redundant predictions arising from potential overlaps among candidate predictions. Here, we uncover its inherent weaknesses (\ie factorial time complexity) and exploit it to perform \sponge. Our objective is to compress the dimensions of all bboxes, thus reducing the \textit{intersection over union} (IOU) between the candidates. Specifically, we define the following loss for an individual bbox:
\begin{IEEEeqnarray}{rCL}   
\mathcal{L}_{\text{box}} = \frac{1}{|\mathcal{S}_{\text{sel}}|} \sum_{b \in \mathcal{S}_{\text{sel}}}\left(\frac{b^w \cdot b^h}{S}\right)^2,
\end{IEEEeqnarray}
where $b^w$ and $b^h$ are the width and height of the bbox, while $S$ is the size of inputs. Generally, reducing the bbox area can create more space for additional candidates, thereby maximizing the number of FPs to overload the NMS algorithm. This greatly extends the processing time per frame and hampers the real-time capability of the OD system.

Thus, the backdoor task loss $\mathcal{L}_{poi}^{SP}$ for \sponge is:  
\begin{IEEEeqnarray}{rCL}   
\displaystyle 
 \mathcal{L}_{\textnormal{poi}}^{SP} &=-\lambda_{1} \mathcal{L}_{\textnormal{obj}} + \lambda_{2}  \mathcal{L}_{\textnormal{cls}}^{SP} + \lambda_{3} \mathcal{L}_{\textnormal{box}}, 
\end{IEEEeqnarray}
where $\lambda_1$, $\lambda_2$ and $\lambda_3$ are pre-defined hyper-parameters.

\subsubsection{$\blind$ Strategy} 
Considering a significant increase in \textit{False Negatives} (FNs) also impacts the clean loss $\mathcal{L}$, we propose the second strategy: generating FNs. As described in Section~\ref{Sec: Relatedwork}, we focus on a more general-purpose object disappearance attack, where a trigger is designed to facilitate the disappearance of global bboxes. \blind involves consists of two components: (a) objectness loss $\mathcal{L}_{obj}$, and (b) classification loss $\mathcal{L}_{\textnormal{cls}}^{BL}$.

\noindent\textbf{Objectness loss.} 
Similarly, to conduct \blind, we reemphasize the critical step in deep learning-based detectors: distinguishing whether regions in the input image are background elements or objects of interest. Hence, \blind ensures that detector $F$ recognizes all regions in poisoned images as background. The objective is to minimize Eq.~(\ref{Eq:Background}).

\begin{table*}[h]
  \centering
  \setlength{\tabcolsep}{2pt}
  \small
  \resizebox{\textwidth}{!}{
    \begin{tabular}{cccllp{1.5cm}llp{1.5cm}llp{1.5cm}lll}
    \toprule[1.5pt]
    \multicolumn{3}{c}{\multirow{2}[4]{*}{Dataset}} & \multicolumn{6}{c}{MS-COCO \cite{lin2014microsoft}}                  & \multicolumn{6}{c}{VOC \cite{everingham2010pascal}} \\
\cmidrule{4-15}    \multicolumn{3}{c}{}  & \multicolumn{3}{c}{Benign~$\uparrow$} & \multicolumn{3}{c}{Poisoned~$\downarrow$} & \multicolumn{3}{c}{Benign~$\uparrow$} & \multicolumn{3}{c}{Poisoned~$\downarrow$} \\
    \midrule
    Detector & Setting & \diagbox{Method}{Metric} & $mAP_{50}$ & $mAP_{75}$ & $mAP$ & $mAP_{50}$ & $mAP_{75}$ & $mAP$ & $mAP_{50}$ & $mAP_{75}$ & $mAP$ & $mAP_{50}$ & $mAP_{75}$ & $mAP$ \\
    \midrule
    \multirow{10}[6]{*}{\makecell{FR}} & No attack & \textit{—} &    58.1   &   40.4    &   37.4    &   53.8    &    35.7   &  33.2     &   76.5    &  55.1     & 49.8      &   73.6    &  51.0     & 46.6 \\
\cmidrule{2-15}          & \multirow{6}[2]{*}{Appearing attack} & \ct{OGA}   &   57.2    &   39.7    &  36.2     &   33.8     &   21.5   &   17.0    &  76.1     &   54.2    &     48.8   &    52.9   & 35.1     &  32.5\\
          &       & \ct{RMA}   &   55.0    &  38.0     &    35.1   &     45.1  &  29.9     &   25.8    &    75.7   &   54.0    &   47.3    &    61.1   &   41.7    & 36.0 \\
          &       & \ct{GMA}   &   52.6    &   36.7    &   34.9    &   34.0    &    17.6   &   14.4    &     74.9  &     53.6  &  46.1     &   49.4    &  32.8     & 28.3 \\
          &       & \ct{UT} &     55.5  &  38.4     &   35.8    & 20.5      & 12.6      &     11.7  &   71.9    &  52.5     & 45.4      &  28.3     &  18.4     &  18.5\\
          &       & \ct{Clean-label} &  56.7     & 39.2      &  36.1     &  48.5     &     32.4  &  29.8     &    76.2   &    54.4   &   48.9    &   67.7    &  45.0     & 41.9 \\
          &       & \cellcolor[rgb]{.95,.95,.95} \sponge \textit{(ours)} & \cellcolor[rgb]{.95,.95,.95}56.1 & \cellcolor[rgb]{.95,.95,.95}38.2 & \cellcolor[rgb]{.95,.95,.95}35.8 & \cellcolor[rgb]{.95,.95,.95}\textbf{0.8} & \cellcolor[rgb]{.95,.95,.95}\textbf{0.5} & \cellcolor[rgb]{.95,.95,.95}\textbf{0.4} & \cellcolor[rgb]{.95,.95,.95}74.5 & \cellcolor[rgb]{.95,.95,.95}54.2 & \cellcolor[rgb]{.95,.95,.95}47.9 & \cellcolor[rgb]{.95,.95,.95}\textbf{1.1} & \cellcolor[rgb]{.95,.95,.95}\textbf{0.6} & \cellcolor[rgb]{.95,.95,.95}\textbf{0.6}\\
\cmidrule{2-15}          & \multirow{3}[2]{*}{Disappearing attack} & \ct{UT} &     54.8  &    37.9   &  35.5     &  26.1     &  15.8     & 16.6      & 72.2      &   53.3    &   47.0    &   27.9    & 19.5      & 17.4 \\
          &       & \ct{Clean-label} &   56.4    &  38.9     & 36.0      & 49.0      & 32.7      &   29.3    &   75.8    &  54.0     &   47.7    &    64.2   &  43.3     & 40.0 \\
          &      & \cellcolor[rgb]{.95,.95,.95} \blind \textit{(ours)} & \cellcolor[rgb]{.95,.95,.95}56.5 & \cellcolor[rgb]{.95,.95,.95}38.7 & \cellcolor[rgb]{.95,.95,.95}36.0 & \cellcolor[rgb]{.95,.95,.95}\textbf{14.2} & \cellcolor[rgb]{.95,.95,.95}\textbf{5.7} & \cellcolor[rgb]{.95,.95,.95}\textbf{6.6} & \cellcolor[rgb]{.95,.95,.95}76.0 & \cellcolor[rgb]{.95,.95,.95}54.0 & \cellcolor[rgb]{.95,.95,.95}48.2 & \cellcolor[rgb]{.95,.95,.95}\textbf{18.5} & \cellcolor[rgb]{.95,.95,.95}\textbf{10.7} & \cellcolor[rgb]{.95,.95,.95}\textbf{11.1} \\
    \midrule  \midrule
    \multirow{10}[6]{*}{\makecell{Y5}} & No attack & \textit{—} &  53.7     &  40.8     &  35.3     & 50.1      &   37.9    &  32.6     &  78.8     & 58.4      &   52.7    & 77.2      & 55.6      &  50.3\\
\cmidrule{2-15}          & \multirow{6}[2]{*}{Appearing attack} & \ct{OGA}   &   52.4    &    38.6   &    34.7   &   30.8    &   22.0    &  18.5     &  77.1     &  56.7     &    50.8   &   56.2    &   39.6    & 35.4 \\
          &       & \ct{RMA}   &    51.6   &  37.7     &  33.0     &  33.1     &  23.6     & 19.7      &    75.9   &  56.2     & 49.7      &  59.9     &  43.7     &37.5  \\
          &       & \ct{GMA}   &   52.4    &   37.9    &  33.5     &   23.5    &   16.0    &    14.2   &    76.4   &   56.5    & 50.1      &   35.2    &   22.3    & 20.8 \\
          &       & \ct{UT} &    52.2   & 38.1      &   34.2    &   23.8    &  17.6     &  15.2     &      75.8 & 56.0      &  49.7     & 42.8      & 29.3      &  26.7\\
          &       & \ct{Clean-label} &     53.6  &   39.5    &  34.8     &  49.0     &  36.5     &   30.0    &  77.3     & 57.0      & 50.8      &  68.8     &47.9       &43.5  \\
          &      & \cellcolor[rgb]{.95,.95,.95} \sponge \textit{(ours)} & \cellcolor[rgb]{.95,.95,.95}52.3 & \cellcolor[rgb]{.95,.95,.95}38.5 & \cellcolor[rgb]{.95,.95,.95}34.5 & \cellcolor[rgb]{.95,.95,.95}\textbf{3.7} & \cellcolor[rgb]{.95,.95,.95}\textbf{2.2} & \cellcolor[rgb]{.95,.95,.95}\textbf{2.1} & \cellcolor[rgb]{.95,.95,.95}77.0 & \cellcolor[rgb]{.95,.95,.95}56.7 & \cellcolor[rgb]{.95,.95,.95}50.4 & \cellcolor[rgb]{.95,.95,.95}\textbf{5.1} & \cellcolor[rgb]{.95,.95,.95}\textbf{3.5} & \cellcolor[rgb]{.95,.95,.95}\textbf{3.6} \\
\cmidrule{2-15}          & \multirow{3}[2]{*}{Disappearing attack} & \ct{UT} &  49.7     &   37.1    &  33.9     &  35.3     &  23.4     &    22.6   &  74.5     &     55.3  &    48.7   &    52.5   &36.1       &33.0  \\
          &       & \ct{Clean-label} & 53.2      &  40.4     &   35.1    &  49.4     & 36.6      & 30.2      &  78.0     &  57.2     &    51.6   &     70.3  &    50.9   &  45.7\\
          &      & \cellcolor[rgb]{.95,.95,.95} \blind \textit{(ours)} & \cellcolor[rgb]{.95,.95,.95}52.4 & \cellcolor[rgb]{.95,.95,.95}38.5 & \cellcolor[rgb]{.95,.95,.95}34.1 & \cellcolor[rgb]{.95,.95,.95}\textbf{17.0} & \cellcolor[rgb]{.95,.95,.95}\textbf{8.3} & \cellcolor[rgb]{.95,.95,.95}\textbf{8.5} & \cellcolor[rgb]{.95,.95,.95}76.8 & \cellcolor[rgb]{.95,.95,.95}56.6 & \cellcolor[rgb]{.95,.95,.95}50.1 & \cellcolor[rgb]{.95,.95,.95}\textbf{28.4} & \cellcolor[rgb]{.95,.95,.95}\textbf{16.9} & \cellcolor[rgb]{.95,.95,.95}\textbf{13.5}\\
    \bottomrule[1.5pt]
    \end{tabular}%
    }
  \caption{Comparison of the performace between SOTA backdoors and ours. For each adversarial case, the best results are highlighted in bold.}
  \label{map}%
\end{table*}%

\noindent\textbf{Classification loss.}
Equally important is reducing the maximum class probability of bboxes to diminish their confidence scores, thereby augmenting the likelihood of their exclusion during the filtering process. In object class predictions, a high concentration of prediction probability signifies model confidence, with the worst performance akin to random guessing (\eg $\sim$10\% accuracy in a 10-class dataset). Consequently, \blind seeks to undermine this confidence. Definition~\ref{Def:2} ensures dispersible predictions.

\begin{definition}
\label{Def:2}
(Mean Prediction Dispersion). Define $\boldsymbol{P}^{(c)}$ as the probability vector for objects of ground-truth class $c$, with each $j$-th entry of $\boldsymbol{P}^{(c)}$ is
\begin{IEEEeqnarray}{rCL} 
P_j^{(c)} \triangleq \frac{\sum_{i=1}^N \mathbb{I}\left\{F_{\boldsymbol{\theta}}\left(\boldsymbol{x}_i\right)=j\right\} \cdot \mathbb{I}\left\{\boldsymbol{y}_i=c\right\}}{\sum_{i=1}^N \mathbb{I}\left\{\boldsymbol{y}_i=c\right\}} .
\end{IEEEeqnarray}
The mean prediction dispersion $\mathcal{R}$ is expressed as
\begin{IEEEeqnarray}{rCL} 
\mathcal{R} \triangleq \frac{1}{N} \sum_{c=1}^C \sum_{i=1}^N \mathbb{I}\left\{\boldsymbol{y}_i=c\right\} \cdot \mathcal{H}\left[\boldsymbol{P}^{(c)}\right],
\end{IEEEeqnarray}
where $\mathcal{H}(\cdot)$ is the entropy~\cite{kullback1997information}.
\end{definition}

Generally speaking, $\mathcal{R}$ measures the prediction dispersion for objects with identical labels; a higher $\mathcal{R}$ indicates less certainty in the detector's output (thus a lower confidence score). To execute \blind, optimizing the mean prediction dispersion is essential. Since it is non-differentiable, we in turn employ differentiable surrogate measures, focusing on maximizing the mean object-wise prediction dispersion to achieve a similar effect, as follows:
\begin{IEEEeqnarray}{rCL} 
\mathcal{L}_{\textnormal{cls}}^{BL}=\frac{1}{N}\sum_{i=1}^N\mathcal{H}\left[\boldsymbol{P}_i\left(F_{\boldsymbol{\theta}}\left(\boldsymbol{x}_i\right)\right)\right].
\end{IEEEeqnarray} 
This method effectively impairs the detector's ability to recognize any objects in the input image while also preventing FP outcomes. Finally, the backdoor task loss for \blind is: $\mathcal{L}_{\textnormal{poi}}^{BL} =\lambda_{1} \mathcal{L}_{\textnormal{obj}} - \lambda_{2} \mathcal{L}_{\textnormal{cls}}^{BL}$.

\section{Experiments}
\subsection{Experimental Settings}
\label{Sec:Exp}

\noindent\textbf{Datasets and detectors.} 
We select \underline{MS-COCO} 2014~\cite{lin2014microsoft} and \underline{PASCAL VOC} (VOC) 07\&12 ~\cite{everingham2010pascal} for evaluation. The representative detectors we choose are one-stage \underline{YOLOv5-s} (Y5) with the CSPDarknet-53 feature extractor and two-stage \underline{Faster R-CNN} (FR)~\cite{ren2015faster} with the ResNet-50 backbone.

\noindent\textbf{Evaluation metrics.} 
We employ the metrics \underline{$mAP_{50}$}, \underline{$mAP_{75}$}, and \underline{$mAP_{50:95}$} (the average $mAP$ values with IOU thresholds varying from 0.5 to 0.95) for comprehensive performance assessment, defaulting to the widely used $mAP_{50:95}$ (denoted as $mAP$ for short). Triggering Success Rate (\underline{TSR}) gauges backdoor effectiveness in the physical world. We also adopt \underline{Latency} and Frames Per Second (\underline{FPS}) to assess the processing efficiency of detectors.

\noindent\textbf{Competitors.}
Given that non-universal methods are not applicable to our attack scenarios, we focus on universal backdoors in Tab.~\ref{Tab:Comparison} for comparison.
Current SOTA backdoor attacks include \ct{OGA} \& \ct{RMA} \& \ct{GMA} \cite{chan2022baddet}, \ct{UT}~\cite{luo2023untargeted} (we adaptively design UT to create numerous false bboxes of random categories and sizes around the trigger), and \ct{Clean-label}~\cite{chen2022clean}.

\noindent\textbf{Attack setting.} 
we employ four commonly used trigger patterns, \ie Chesssboard, Kitty, Basketball, and Random trigger (see Fig.~\ref{Fig:trigger}(a)), with the Chesssboard as our default. 
Notably, for a fair comparison, each poisoned test sample features only one randomly placed trigger. We set the default trigger size as $\frac{1}{64}$ of the whole image area (i.e., $\frac{1}{8}$ width and $\frac{1}{8}$ height). The data poisoning rate is 10\%, while the trigger ratio $\eta$ is 0.5. We set $\lambda_1=0.6$, $\lambda_2=0.3$, and $\lambda_3=0.1$.

\subsection{Comparison with State-of-the-Art}   
We train backdoor detectors on MS-COCO and VOC using the methods mentioned in Section~\ref{Sec:Exp} and evaluate their mAP metrics on both benign and poisoned samples. From Tab.~\ref{map}, the mAP values for most methods on benign samples, including our DC, do not exhibit apparent declines. Backdoor attacks like \ct{OGA} fail to achieve superior performance as they only cause localized errors, leaving the overall detection results largely unaffected. \ct{RMA} and \ct{Clean-label} are further constrained in effectiveness as they necessitate the victim to be in close proximity to the trigger. The performance of \ct{GMA} performs relatively better by causing global misclassification, yet its effectiveness is constrained by lower attack success rates. The overall performance of \ct{UT} is commendable, reflected in the significant reduction of the mAP in poisoned samples. However, it is also notably inferior to our \ourmethod, which consistently achieves optimal attack performance (in line with our Definition~\ref{Def:1}) across various detectors, datasets, and settings. Specifically, the \sponge can reduce the detectors' mAP values on poisoned samples nearly to 0, while the \blind can decrease the mAP value to less than 15\% (\ie most ground-truth objects are missed). In addition, we tested \sponge's system-level disruptive impact (see Tab.~\ref{tab:latency}). On average, \sponge introduced a delay of $\sim$7-10$\times$ in the processing time of a single poisoned image, posing a significant threat to real-time critical detection systems.

\begin{table}[t]
  \centering
  \setlength{\tabcolsep}{2pt}
  \small
  \resizebox{0.48\textwidth}{!}{
    \begin{tabular}{cccp{1.8cm}p{1.8cm}p{1.8cm}p{1.8cm}}
    \toprule[1.5pt]
    \multirow{2}[2]{*}{Dataset} & \multirow{2}[2]{*}{Detector} & \multirow{2}[4]{*}{Method} & \multicolumn{4}{c}{Latency (ms)~\bm{$\downarrow$} / FPS (f/s)~\bm{$\uparrow$}} \\
    \cmidrule{4-7}          &       &       & \multicolumn{1}{c}{CPU} & \multicolumn{1}{c}{RTX 3090} & \multicolumn{1}{c}{RTX 4090} & \multicolumn{1}{c}{Avg.} \\
    \midrule
    \multirow{4}[3]{*}{MS-COCO} & \multirow{2}[2]{*}{Y5} & No attack & 99.3~/~{10.1} & 6.3~/~{158.7} & 5.1~/~{196.1} & 36.9~/~{121.6} \\ 
    \addlinespace[3pt]
          &       & \cellcolor[rgb]{.95,.95,.95}\sponge & \cellcolor[rgb]{.95,.95,.95}691.2~/~1.4 & \cellcolor[rgb]{.95,.95,.95}137.5~/~7.3 & \cellcolor[rgb]{.95,.95,.95}114.3~/~8.7 &  \cellcolor[rgb]{.95,.95,.95}314.3~/~17.4 \\
    \cmidrule{2-7}          & \multirow{2}[1]{*}{FR} & No attack & 1937.5~/~0.5 & 94.4~/~10.6 & 77.0~/~12.9 & 702.9~/~8.0 \\
    \addlinespace[3pt]
          &       & \cellcolor[rgb]{.95,.95,.95}\sponge & \cellcolor[rgb]{.95,.95,.95}20410.1~/~0.1 & \cellcolor[rgb]{.95,.95,.95}864.2~/~1.2 & \cellcolor[rgb]{.95,.95,.95}753.8~/~1.3 & \cellcolor[rgb]{.95,.95,.95}7342.7~/~0.9 \\
    \midrule[1.5pt]
    \multirow{4}[2]{*}{VOC} & \multirow{2}[1]{*}{Y5} & No attack & 90.8~/~11.0 & 6.1~/~163.9 & 4.9~/~204.1 & 33.9 ~/~126.3 \\
    \addlinespace[3pt]
          &       & \cellcolor[rgb]{.95,.95,.95}\sponge & \cellcolor[rgb]{.95,.95,.95}620.7~/~1.6 & \cellcolor[rgb]{.95,.95,.95}120.9~/~8.3 & \cellcolor[rgb]{.95,.95,.95}102.1~/~9.8 & \cellcolor[rgb]{.95,.95,.95}281.2~/~6.6 \\
    \cmidrule{2-7}          & \multirow{2}[1]{*}{FR} & No attack & 1880.2~/~0.5 & 90.5~/~11.1 & 74.7~/~13.4 & 681.8~/~8.3 \\
    \addlinespace[3pt]
          &       & \cellcolor[rgb]{.95,.95,.95}\sponge& \cellcolor[rgb]{.95,.95,.95}17596.4~/~0.1 & \cellcolor[rgb]{.95,.95,.95}825.2~/~1.2 & \cellcolor[rgb]{.95,.95,.95}740.3~/~1.4 & \cellcolor[rgb]{.95,.95,.95}6387.3~/~0.9 \\
    \bottomrule[1.5pt]
    \end{tabular}%
  }
  \caption{Average latency for processing a poisoned image and FPS on various devices.}
  \label{tab:latency}%
\end{table}%

\subsection{The Resistance to Potential Backdoor Defenses}
This section evaluates \ourmethod against two representative defenses, \ie fine-tuning~\cite{sha2022finetuning,zhou2024genaf,zhang2024does} and pruning~\cite{liu2018fine,zhou2023advclip,zhou2023advencoder}, on Y5 + MS-COCO. 



\begin{figure}[!t]   \centering   
\subfigure[Fine-tuning]{\includegraphics[width=0.49\linewidth]{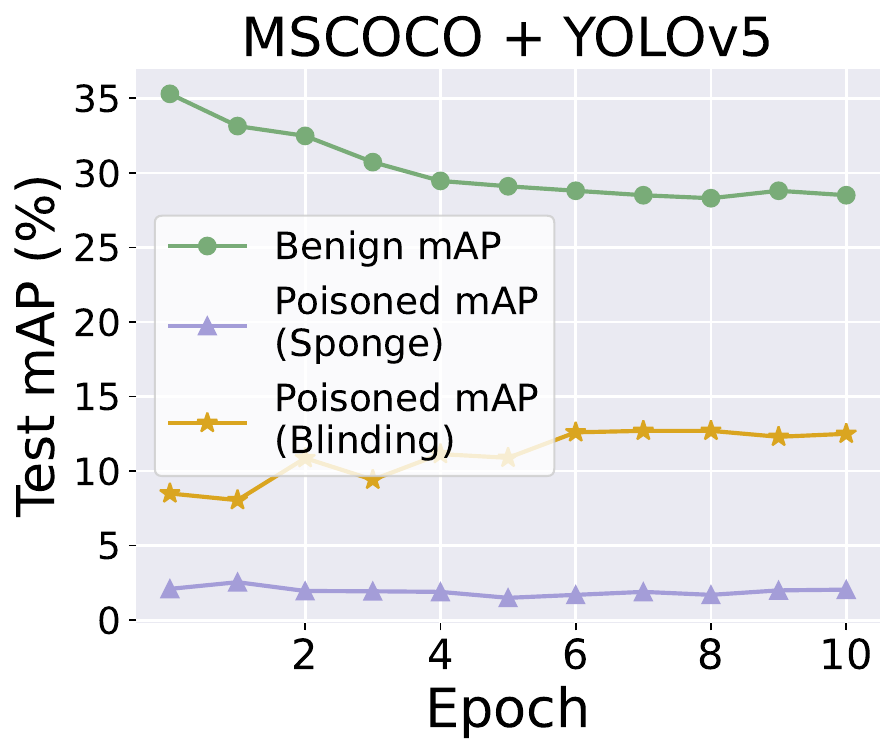}}       
\subfigure[Pruning]{\includegraphics[width=0.49\linewidth]{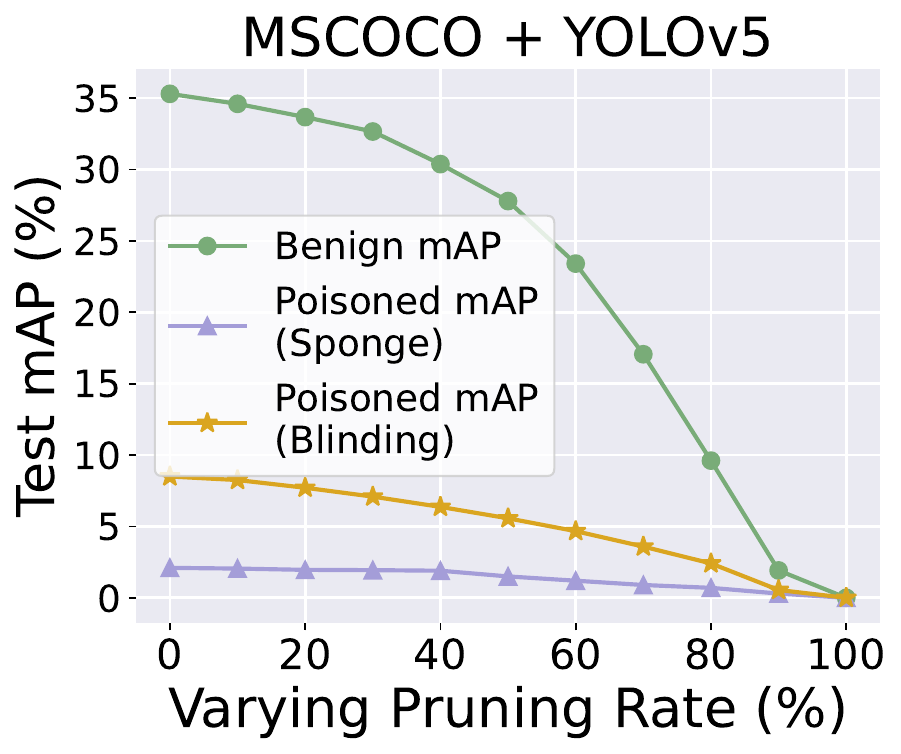}}            \caption{The effectiveness of several potential defenses}    \label{fig:defense}   
\end{figure}

\noindent\textbf{Fine-tuning.}
We fine-tune the detector with $10\%$ benign testing samples, maintaining the original learning rate. Fig.~\ref{fig:defense}(b) shows the resistance of our method to fine-tuning, with the poisoned mAP remaining under $13\%$ after fine-tuning. The limited trigger features in the fine-tuning dataset result in only partial correction of the backdoored model, and the samples are inadequate to eliminate the backdoor fully.

\noindent\textbf{Pruning.}
Following the classical settings~\cite{liu2018fine,chen2022clean}, we assess the backdoor model using the clean test samples and arrange neurons in ascending order by their average activation values. Then we prune these neurons in order. As shown in Fig.~\ref{fig:defense}(c), increasing the pruning rate does not restore the poisoned mAP. This suggests an overlap between backdoor and clean neurons, hindering the pruning strategy from effectively segregating them.

\begin{figure*}[t]  
\centering     
\includegraphics[width=\linewidth]{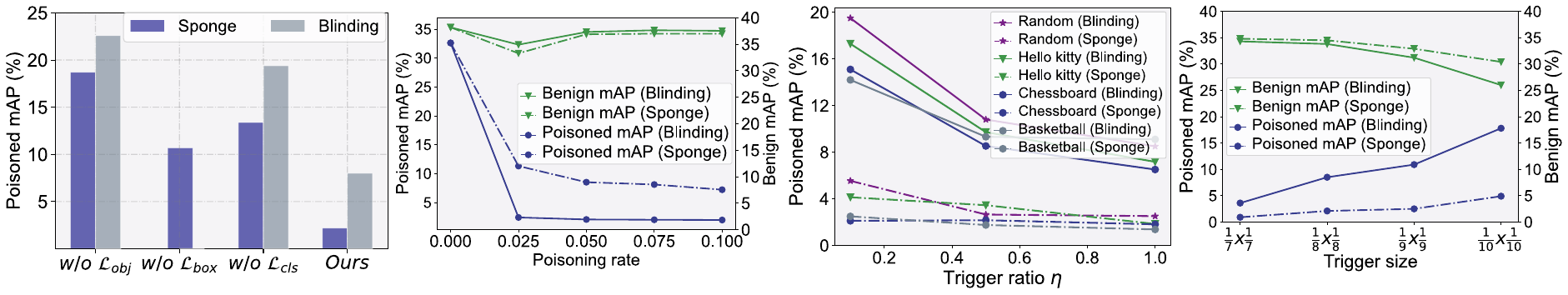}
\caption{Impact of loss functions, parameter and trigger variations on the effect of our \ourmethod} 
\label{Fig:Ablation}
\end{figure*}

\begin{figure}[!h] 
\centering       
\includegraphics[width=0.48\textwidth]{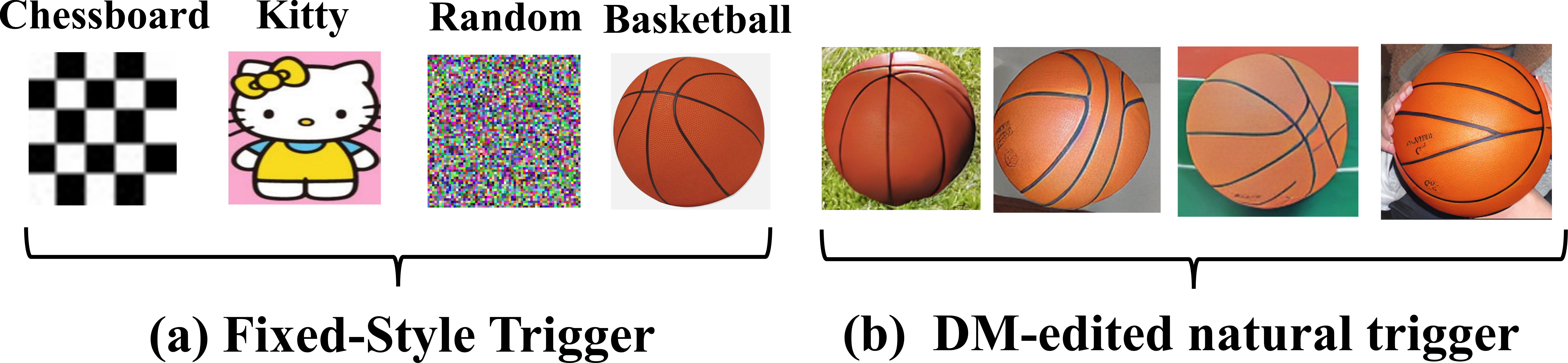} \caption{Experimental triggers details}       \label{Fig:trigger}   
\end{figure}

\subsection{Abaltion Study for \ourmethod} 
We explore the effect of poisoning rate, trigger size, trigger ratio $\eta$, trigger patterns, and different modules. Using Y5 + MS-COCO, we assess our attacks, maintaining consistency with the parameters used in Tab.~\ref{map}. Each study varies one single parameter to isolate its impact. Fig.~\ref{Fig:Ablation} leads to key findings: 1) the poisoning rate slightly impacts the attack effect; 2) a larger trigger size contributes to better attack performance; 3) a higher trigger ratio ($\eta$) marginally impacts the poisoned mAP. Also, four triggers show similar results, demonstrating the generalizability of using various triggers.

\paragraph{The Effect of $\mathcal{L}_{obj}$ \&  $\mathcal{L}_{box}$ \& $\mathcal{L}_{cls}$.} 
Fig.~\ref{Fig:Ablation} underscores the necessity of all three loss designs for achieving high attack efficacy. Notably,  $\mathcal{L}_{\textnormal{obj}}$ emerges as the most critical element for enhancing attack impact. For \sponge, without $\mathcal{L}_{\textnormal{box}}$ and $\mathcal{L}_{\textnormal{cls}}$, the poisoned map showed improvements of approximately 9\% and 12\%, respectively. Hence, these results highlight the importance of integrating all three losses. For \blind, the combination of $\mathcal{L}_{\textnormal{obj}}$ and $\mathcal{L}_{\textnormal{cls}}$ is also of great significance.


\begin{figure}[t]  
\centering \includegraphics[width=0.47\textwidth]{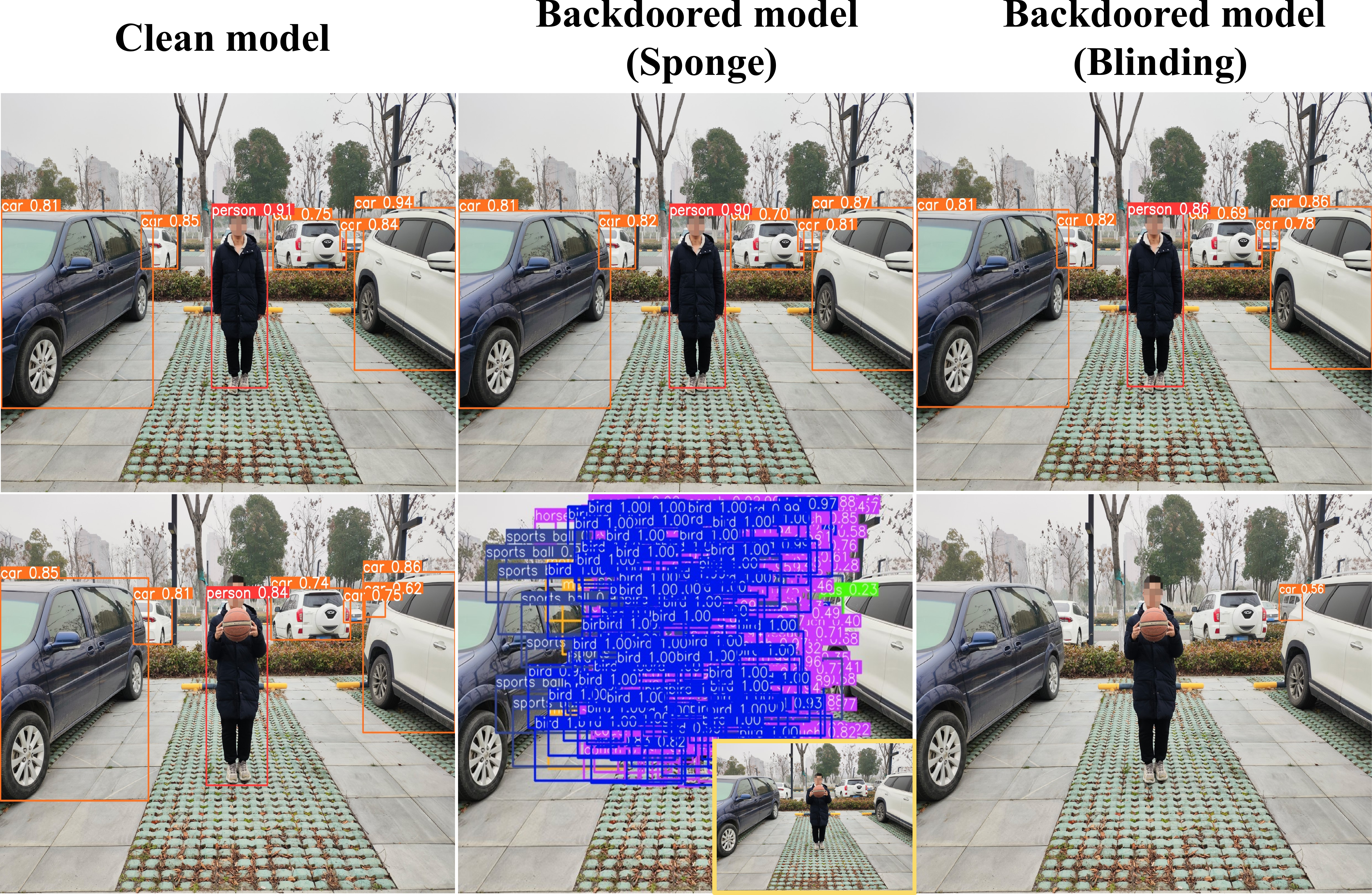}        \caption{Our \ourmethod in the physical world using Y5 + MS-COCO. The first row shows the detection results of benign samples, while the second row presents the results of backdoored samples, serving a basketball in the physical world as the trigger. All images were captured in secure environments with sensitive information obscured.}  \label{Fig:Physical_world}    \end{figure}   

\begin{figure}[!t] 
\centering          \includegraphics[width=0.47\textwidth]{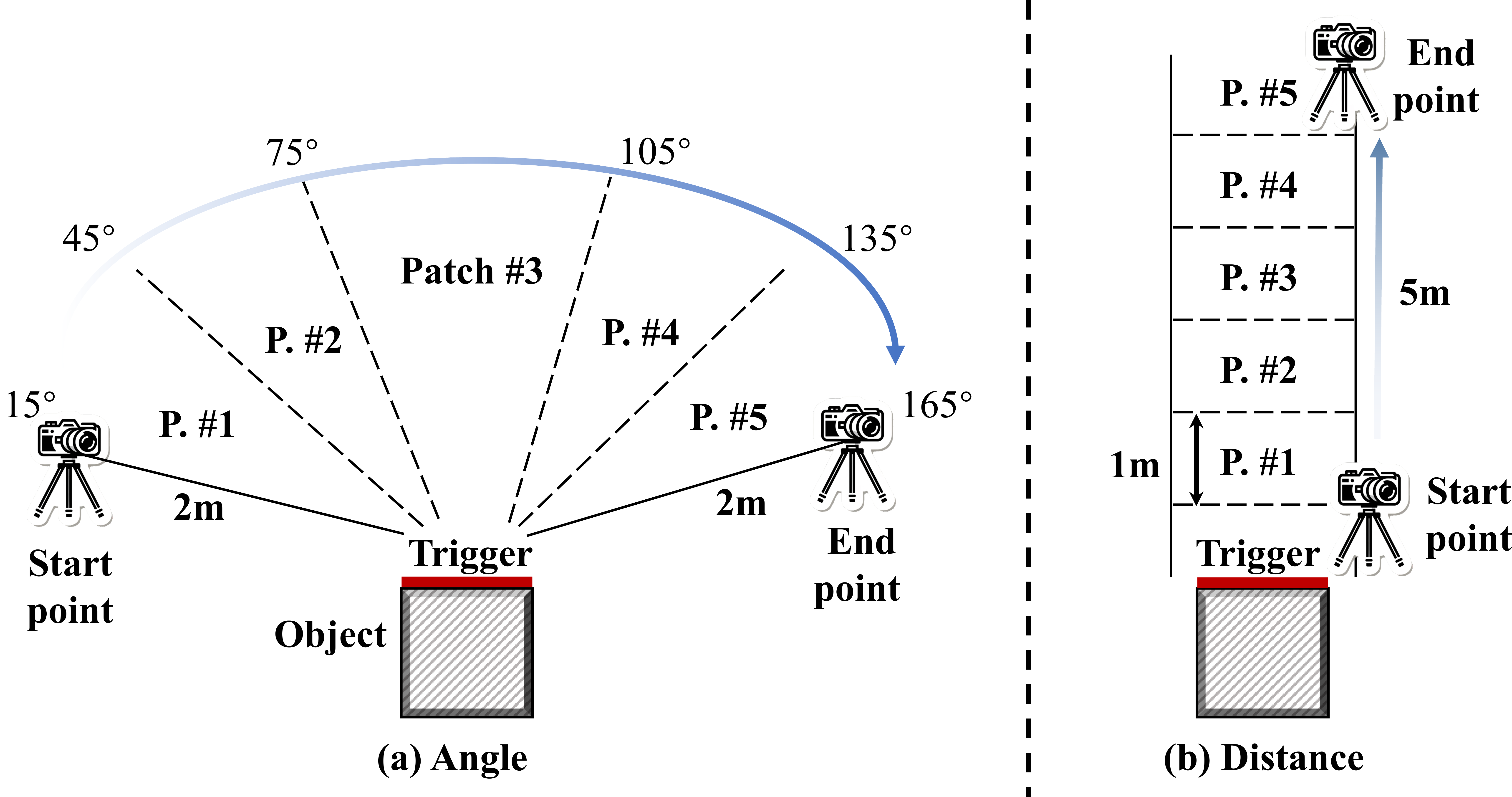}      \caption{Sampling in physical world. We systematically vary the angle and distance to mirror real-world conditions more accurately.}       \label{Fig:phy-sample} 
\end{figure}

\begin{table}[t]    
\centering   
\resizebox{0.48\textwidth}{!}{     
\begin{tabular}{cccccccc}    
\toprule[1.5pt]     
\multirow{2}[4]{*}{Variable} & \multirow{2}[4]{*}{Trigger} & \multicolumn{6}{c}{TSR~(\%) $\uparrow$} \\ \cmidrule{3-8}          &       & P. \#1 & P. \#2 & P. \#3 & P. \#4 & P. \#5 & Avg. \\     \midrule     \multirow{5}[2]{*}{Angle} & Kitty & 14.68 & 64.29 & 92.94 & 61.90 & 14.76 & 49.71 \\           & Chessboard & 16.62 & 67.45 & 93.08 & 65.59 & 15.40 & 51.63 \\           & Basketball  & 15.01 & 54.83 & \textbf{94.95} & 60.27 & 14.29 & 47.87 \\           & Random & 4.69 & 18.11 & 88.41 & 16.80 & 5.67 & 26.74 \\           & \cellcolor[rgb]{.95,.95,.95}Natural Basketball & \cellcolor[rgb]{.95,.95,.95}\textbf{58.45} & \cellcolor[rgb]{.95,.95,.95}\textbf{84.96} & \cellcolor[rgb]{.95,.95,.95}91.53 & \cellcolor[rgb]{.95,.95,.95}\textbf{86.32} & \cellcolor[rgb]{.95,.95,.95}\textbf{60.03} & \cellcolor[rgb]{.95,.95,.95}\textbf{76.26} \\     \midrule[1.5pt]     \multirow{5}[2]{*}{Distance} & Kitty & 52.43 & 84.26 & 24.10 & 3.93 & 0.00 & 32.94 \\           & Chessboard & 54.09 & 82.79 & 31.72 & 4.86 & 0.00 & 34.69 \\           & Basketball  & 53.88 & 90.10 & 39.60 & 4.57 & 0.00 & 37.63 \\           & Random & 61.30 & 72.22 & 2.87 & 0.00 & 0.00 & 27.28 \\           & \cellcolor[rgb]{.95,.95,.95}Natural Basketball & \cellcolor[rgb]{.95,.95,.95}\textbf{53.69} & \cellcolor[rgb]{.95,.95,.95}\textbf{96.01} & \cellcolor[rgb]{.95,.95,.95}\textbf{98.01} & \cellcolor[rgb]{.95,.95,.95}\textbf{86.67} & \cellcolor[rgb]{.95,.95,.95}\textbf{40.50} & \cellcolor[rgb]{.95,.95,.95}\textbf{74.98} \\     \bottomrule[1.5pt]     \end{tabular} 
}
\caption{The backdoor's TSR of Y5 + MS-COCO in the physical world across various patches (refer to Fig.~\ref{Fig:phy-sample})}
\label{tab:TSR} 
\end{table}%

\subsection{Results in Physical World} 
In real-world settings, the effectiveness of physical triggers is predominantly affected by universal factors such as distance and angles~\cite{qian2023robust,wenger2021backdoor}. Therefore, we investigate into these two variables as illustrated in Fig.~\ref{Fig:phy-sample} by capturing 100 consecutive frames in each region while varying the angle from 15° to 165° and the distance from 1m to 5m between the camera and the physical trigger. Tab.~\ref{tab:TSR} presents \ourmethod's TSR values with various triggers, calculated as the fraction of successful attacks out of 100 frames. The results indicate that backdoor attacks employing triggers of a fixed pattern exhibit unstable effects, with their TSR significantly decreasing as the angle and distance change.

We note that the backdoor's poor robustness in the physical world stems from its reliance on a fixed trigger pattern. Hence, we propose adopting semantic features (\eg characteristics of a basketball) to enable the learning of more flexible trigger patterns during training. The labor-intensive task of collecting natural objects as triggers is circumvented by pre-trained \textit{diffusion models} (DM)~\cite{yang2023paint}, which use triggers as references to contextually edit clean images, creating subtly varied triggers (see Fig.~\ref{Fig:trigger}(b)). 
Specifically, under identical settings, we shift to using this dynamic triggers for training the backdoor model, and utilize natural objects (\ie a real basketball in the physical world) for activation during testing. According to Tab.~\ref{tab:TSR}, the \textit{Natural Basketball}'s TSR remains high despite variations in angles and distances, affirming the efficacy of this poisoning strategy. Finally, armed with this new strategy, we successfully liberated \ourmethod from reliance on fixed trigger patterns, extending it to the physical world. Fig.~\ref{Fig:Physical_world} showcases a demonstration of backdoor activation in the physical world.

\section{Conclusions}
In this paper, we introduce \ourmethod, a novel backdoor paradigm for OD with two unique strategies. Notably, we are the first to present a sponge attack via backdoors, opening a new attack surface in OD. Extensive evaluation on both single and two-stage models validate its effectiveness and generalization. Finally, we extend \ourmethod to the physical world, also achieving excellent attack effectiveness.

\section*{Acknowledgments} 
Shengshan’s work is supported in part by the National Natural Science Foundation of China (Grant Nos. 62372196, U20A20177). Shengshan Hu is the corresponding author. 

\bibliographystyle{named}
\bibliography{ijcai24}

\end{document}